\documentclass{article}
\usepackage{spconf,amsmath,graphicx}

\usepackage{amssymb}
\usepackage{xcolor}
\usepackage{url}
\usepackage{multirow}
\usepackage{caption}
\usepackage{subcaption}
\usepackage{hyperref}

\DeclareMathOperator{\argmax}{arg\,max}

\title{Self-Supervised Syllable Discovery\\Based on Speaker-Disentangled HuBERT}
\twoauthors
  {Ryota Komatsu}{Independent Researcher}
  {Takahiro Shinozaki}{Tokyo Institute of Technology}

\begin{document}
%
\maketitle
\begin{abstract}
Self-supervised speech representation learning has become essential for extracting meaningful features from untranscribed audio. Recent advances highlight the potential of deriving discrete symbols from the features correlated with linguistic units, which enables text-less training across diverse tasks. In particular, sentence-level Self-Distillation of the pretrained HuBERT (SD-HuBERT) induces syllabic structures within latent speech frame representations extracted from an intermediate Transformer layer. In SD-HuBERT, sentence-level representation is accumulated from speech frame features through self-attention layers using a special CLS token. However, we observe that the information aggregated in the CLS token correlates more with speaker identity than with linguistic content. To address this, we propose a speech-only self-supervised fine-tuning approach that separates syllabic units from speaker information. Our method introduces speaker perturbation as data augmentation and adopts a frame-level training objective to prevent the CLS token from aggregating paralinguistic information. Experimental results show that our approach surpasses the current state-of-the-art method in most syllable segmentation and syllabic unit quality metrics on Librispeech, underscoring its effectiveness in promoting syllabic organization within speech-only models\footnote{Codes and models: \url{https://github.com/ryota-komatsu/speaker_disentangled_hubert}}.
\end{abstract}
\section{Introduction}
\label{sec:intro}
Self-supervised speech representation learning has emerged as a key technique as it enables speech models to extract good features from untranscribed audio.
Some of those works have shown that hidden units obtained by discretizing learned features highly correlate with linguistic units, e.g., phones~\cite{9585401}, syllables~\cite{peng23e_interspeech,10446062}, and words~\cite{peng22c_interspeech}.
By utilizing them as pseudo-labels of untranscribed audio, we can train transcript-less models for various spoken language processing tasks, including speech synthesis~\cite{lakhotia-etal-2021-generative,polyak21_interspeech,10158503,zhang-etal-2023-speechgpt}, spoken language understanding~\cite{wu23g_interspeech,fang2024integrating}, speech-to-speech translation~\cite{lee-etal-2022-textless,huang2023transpeech}, and spoken language acquisition~\cite{10096250}.
Among them, with a few exceptions, most works have focused on the shortest phonetic units.
While previous work has shown that a phone-level representation quality highly correlates with the accuracy of automatic speech recognition~\cite{chang23_interspeech}, we often need higher-level information to understand spoken language, where semantic comprehension is essential~\cite{fang2024integrating}.

Recent studies have shown that self-supervised fine-tuning of the pretrained HuBERT~\cite{9585401} naturally induces syllabic organization in an intermediate network layer~\cite{peng23e_interspeech,10446062}.
First, Peng \textit{et al.} demonstrated that syllabic organization emerges in Visually-Grounded HuBERT (VG-HuBERT), which learns the co-occurrence of matched speech-image pairs~\cite{peng23e_interspeech}.
They concluded that visual grounding is responsible for the ability as the same phenomenon does not emerge with the original masked language modeling objective.

Afterwards, Cho \textit{et al.} revealed that syllabic-organized representations also emerge in a speech-only model through sentence-level self-distillation fine-tuning of the pretrained HuBERT, named Self-Distilled HuBERT (SD-HuBERT)~\cite{10446062}.
Like BERT in natural language processing, sentence-level representation was aggregated through self-attention layers using a special learnable token, called the CLS token, concatenated with input speech frame feature sequence~\cite{devlin-etal-2019-bert}.
Additionally, they proposed the spoken sentence ABX (SSABX) task for evaluating sentence discriminability of speech models.
Speech-only models are beneficial in two aspects: 1) no image labels are required, and 2) they can learn concepts without visual entities~\cite{10446062}.
However, their experimental results on the SSABX task have shown that distilled information within the CLS token might be dominated by paralinguistic content.
In fact, we observed a dependence between speaker IDs and the predicted pseudo-categories.
Since syllabic units should be speaker-invariant, this may be a potential factor degrading the quality of linguistic features.

Some works have introduced speaker disentanglement for self-supervised training of speech models~\cite{pmlr-v162-qian22b,chang23_interspeech}.
Chang \textit{et al.} proposed speaker-invariant clustering (Spin), which learns speaker-invariant speech representations by performing swapped prediction between original and speaker-perturbed waveforms~\cite{chang23_interspeech}.
However, the representation obtained by their method is at the phone-level.

\begin{figure*}[t]
    \centering
    \begin{subfigure}{0.33\textwidth}
        \includegraphics[width=\textwidth]{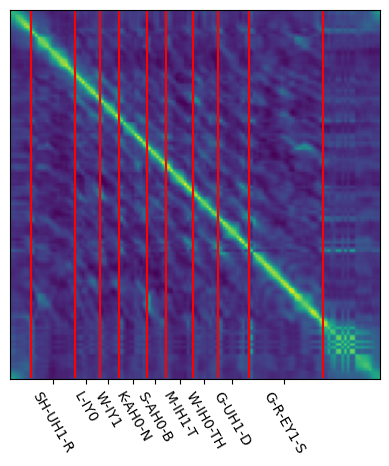}
        \caption{HuBERT $(l=8)$}
        \label{fig:similarity_mat_hubert}
    \end{subfigure}
    \begin{subfigure}{0.33\textwidth}
        \includegraphics[width=\textwidth]{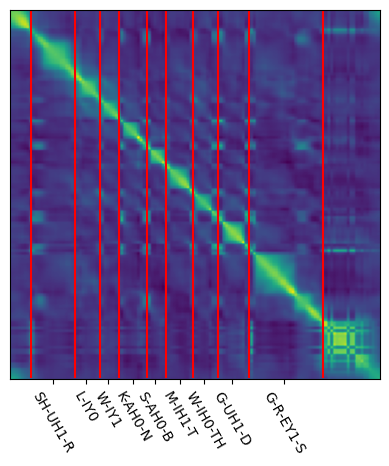}
        \caption{SD-HuBERT $(l=9)$}
        \label{fig:similarity_mat_sdhubert}
    \end{subfigure}
    \begin{subfigure}{0.33\textwidth}
        \includegraphics[width=\textwidth]{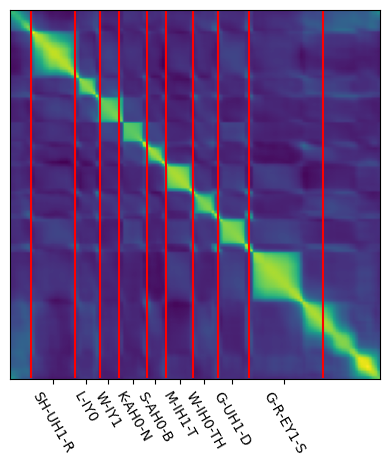}
        \caption{Ours $(l=8)$}
        \label{fig:similarity_mat_ours}
    \end{subfigure}
    \caption{Self-similarity matrices of latent speech frame representations from the $l$-th Transformer layer. Red lines are ground truth syllable boundaries. The speech sample is ``SURELY WE CAN SUBMIT WITH GOOD GRACE''.}
    \label{fig:similarity_mat}
\end{figure*}

In this paper, we propose a teacher-student learning-based speech-only syllabic unit discovery method that integrates speaker information disentanglement for improved performance.
Following~\cite{chang23_interspeech}, to obtain speaker-invariant representations, we constrain a speech model to extract consistent features between the original speech and its speaker-perturbed version.
Moreover, unlike~\cite{10446062}, we avoid using the CLS token and employ a frame-level training objective to prevent the aggregation of paralinguistic information.
Experimental results show that our proposed method outperforms the current state-of-the-art method in most evaluation metrics for syllable segmentation and syllabic unit quality, demonstrating the efficacy of our proposed method.
Finally, our ablation study reveals that the essential factor for syllabic organization is the use of higher Transformer~\cite{NIPS2017_3f5ee243} layers as the student's learning targets that correlate with linguistically coarse-grained units.

\section{Preliminary}
\subsection{Baseline method}
SD-HuBERT~\cite{10446062} is a HuBERT model fine-tuned with self-DIstillation with NO labels (DINO)~\cite{Caron_2021_ICCV}, a self-supervised learning (SSL) method that demonstrated impressive semantic segmentation ability in image processing.
As with DINO, the student and teacher, which have the same architecture but different parameters $\theta_s$ and $\theta_t$, are trained at the sentence-level.
Given a speech frame feature sequence $x_{1:T}$, sentence information is accumulated through self-attention layers using a learnable aggregator token embedding (CLS) concatenated with the input $x_{1:T}$.
The aggregated representation $\mathrm{CLS}^{(L)}$ from the last encoder layer  is fed into a classification head to predict a pseudo-category $c$ using a categorical distribution $p_{\theta}(c\mid\mathrm{CLS}^{(L)})$.
Through self-distillation, syllabic organization naturally emerges in the latent speech frame representations $z_{1:T}^{(l)}$ extracted from the $l$-th Transformer encoder layer of the student.
As shown in Fig.~\ref{fig:similarity_mat_sdhubert}, creating a self-similarity matrix $z_{1:T}^{(l)}{z_{1:T}^{(l)}}^\top\in\mathbb{R}^{T\times T}$ reveals block structures corresponding to syllabic units.
To find the unit boundaries, the minimum cut algorithm~\cite{malioutov-barzilay-2006-minimum,peng23e_interspeech} is used.

\subsection{Analysis}
The predicted sentence-level category for an utterance is obtained by taking the argmax of $p_{\theta_s}(c\mid\mathrm{CLS}^{(L)})$.
In the original paper, experimental results evaluating sentence discriminability suggested that paralinguistic information might dominate the aggregated information.
However, it has not been verified whether this is true, nor has it been determined which types of paralinguistic information predominate.
To investigate the question, we calculated the speaker-normalized mutual information $I(X; Y) / H(X) = 1 - H(X\mid Y) / H(X)$ between speaker ID $X$ and predicted category $Y$, which means the reduction rate of uncertainty about speaker IDs after observing predicted categories~\cite{9585401}.
We observed a very large score of 0.61 on Librispeech~\cite{7178964} test set, which indicates that the aggregator learned to discriminate speaker identity rather than linguistic content.

\section{Proposed Method}
\label{sec:method}

\subsection{Self-supervised fine-tuning}

\begin{figure*}[t]
\centering
\includegraphics[width=0.63\linewidth]{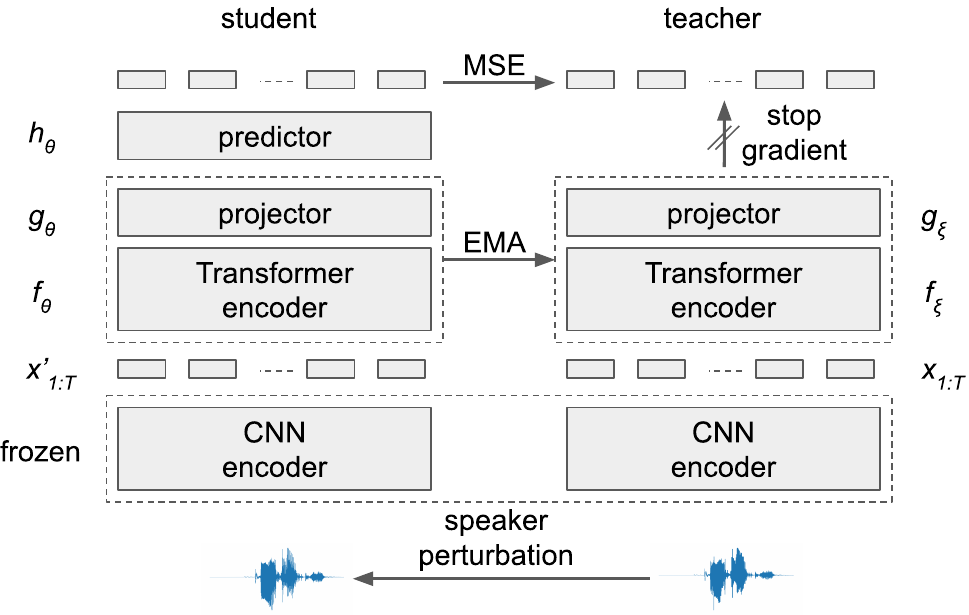}
\caption{The proposed framework.}
\label{fig:model}
\end{figure*}

Fig.~\ref{fig:model} shows the proposed method.
It is based on Bootstrap Your Own Latent (BYOL)~\cite{NEURIPS2020_f3ada80d}, which has been demonstrated superior performance over other SSL methods including DINO in image segmentation task~\cite{Caron_2021_ICCV}.
The model has two branches: student and teacher.
The student consists of a convolutional neural network (CNN) encoder, a Transformer encoder $f_\theta$, a projector $g_\theta$, and a predictor $h_\theta$.
CNN and Transformer encoders are initialized with the pretrained HuBERT.
A teacher parametrized by $\xi$ has the same architecture except that it does not have a predictor.
This additional predictor in the student encodes more information and prevents the model from collapsing into trivial solutions, e.g., outputting the same representations for any inputs~\cite{NEURIPS2020_f3ada80d}.
Prediction targets for the student are the projections of outputs from the last Transformer layer of the teacher (target\_layer=projector).
During training, the stop-gradient is applied to teacher outputs so that gradients backpropagate only to the student.
For efficient training, we freeze the student and teacher CNN encoders.
The remaining teacher parameters are updated with an exponential moving average (EMA) of the student parameters.

To disentangle speaker-dependent information while maintaining linguistic content, we apply the speaker perturbation method proposed in~\cite{NEURIPS2021_87682805} with some modifications.
The original algorithm applies three functions: 1) random formant shift, 2) pitch randomization, and 3) random frequency shaping.
We observe that random perturbations, e.g., doubling the pitch of a female voice, often result in unrealistic speech.
To make perturbed speech realistic, we restrict formant shifts and pitch randomization only to male-to-female and female-to-male conversions.
We calculate the average pitch for each utterance to predict the speaker's gender. If the value is above a threshold, we apply female-to-male conversion; if it is below, we apply male-to-female conversion.
Note that we do not use any speaker gender labels.
An original waveform and the speaker-perturbed one are passed to the teacher and student, respectively.

In SD-HuBERT, minimizing a sentence-level loss led to the aggregation of speaker information in the CLS token representation. Instead, our training objective is the \textit{frame-wise} mean squared error (MSE) between $\ell_2$-normalized student and teacher outputs:
\begin{align*}
\mathcal{L}_{\theta,\xi} &= \frac{1}{T}\sum_{t=1}^T
\left\lVert
\frac{(h_\theta\circ g_\theta\circ f_\theta)(x'_t)}{\lVert(h_\theta\circ g_\theta\circ f_\theta)(x'_t)\rVert}
-
\frac{(g_\xi\circ f_\xi)(x_t)}{\lVert(g_\xi\circ f_\xi)(x_t)\rVert}
\right\rVert^2,
\end{align*}
where $x_t$ and $x'_t$ are the $t$-th speech frames of the original and speaker-perturbed speech, respectively.
Note that minimizing the loss is equivalent to maximizing the cosine similarity between student and teacher outputs.

\subsection{Unit segmentation}

\begin{figure*}[t]
\centering
\includegraphics[width=0.65\linewidth]{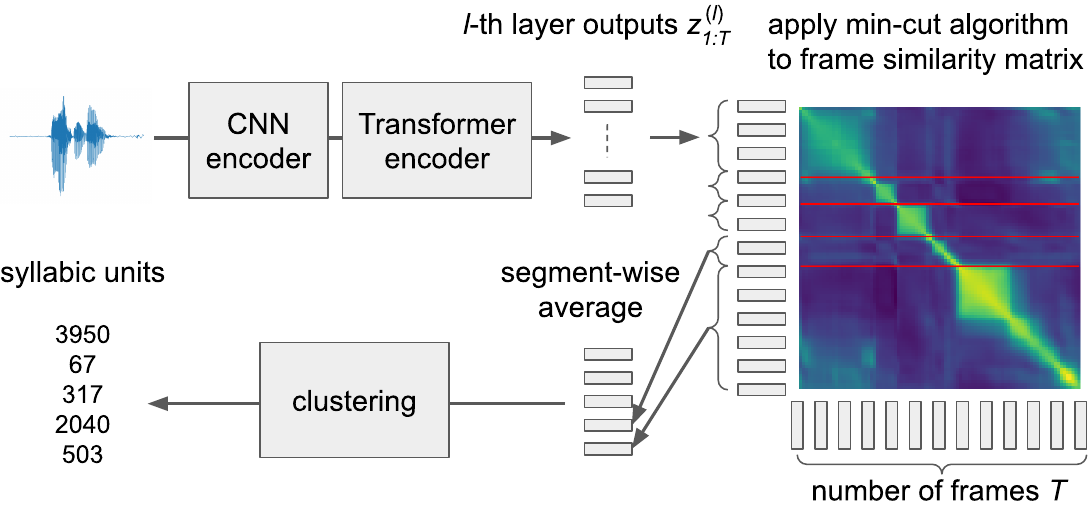}
\caption{Overview of syllabic unit discovery.}
\label{fig:sylseg}
\end{figure*}

Similar to~\cite{10446062}, the proposed method induces syllabic organization in the self-similarity matrices of latent speech frame representations extracted from an intermediate Transformer layer of the student.
We use the syllabic unit discovery algorithm proposed in ~\cite{peng23e_interspeech}, as illustrated in Fig.~\ref{fig:sylseg}.
The procedure consists of three steps: 1) minimum cut algorithm, 2) average pooling within segments, and 3) two-step clustering.
Given the student's $l$-th Transformer layer outputs $z_{1:T}^{(l)}\in\mathbb{R}^{T\times D}$, where $D$ is the hidden size, we firstly calculate the self-similarity matrix $z_{1:T}^{(l)}{z_{1:T}^{(l)}}^\top\in\mathbb{R}^{T\times T}$ based on the dot product similarity measure.
By applying the minimum cut algorithm to the self-similarity matrix, we obtain $S$ syllable segments in time complexity of $O(ST^2)$, where $S$ is a hyperparameter predefined for each utterance.
Then, we average features within each segment.
Finally, $K$-Means followed by agglomerative clustering on $K$ cluster centers assigns pseudo-syllabic units to the segments.

\section{EXPERIMENTAL SETUP}

\subsection{Evaluation metrics}

To evaluate syllable segmentation results, following~\cite{10446062}, we measured precision, recall, F1, and R-value~\cite{rasanen09b_interspeech} of syllable boundaries with a tolerance of up to 50 ms.
Additionally, we evaluate the quality of the discovered syllabic units using syllable purity, cluster purity, and mutual information between ground truth syllables and pseudo-syllabic units~\cite{9585401,peng23e_interspeech,10446062}:
\begin{align*}
\mathrm{syllable\ purity}=&\mathbb{E}_{p(u)}[p(\argmax_s p(s,u)\mid u)],\\
\mathrm{cluster\ purity}=&\mathbb{E}_{p(s)}[p(\argmax_u p(s,u)\mid s)],\\
\mathrm{mutual\ info.}=&\mathbb{E}_{p(s,u)}\left[\log\frac{p(s,u)}{p(s)p(u)}\right],
\end{align*}
where $s$ is a syllable, $u$ is a pseudo-unit, and $p(s,u)$ is the joint distribution between them.
The matching between ground truth and predicted syllables is obtained by solving the maximum weight matching in the temporal intersection-over-union matrix between ground truth and predicted syllable segments.
Moreover, to assess whether speaker information is disentangled by the proposed method, we evaluated accuracy on a speaker identification task.

\subsection{Dataset}
We used Librispeech~\cite{7178964} to train the proposed model, which contained 960 hours of speech data.
For evaluation, we used syllable alignments released in~\cite{10446062}.
In speaker identification analysis, we used VoxCeleb1~\cite{nagrani17_interspeech}.

\subsection{Implementation details}

We used the HuBERT-base architecture, which has 12 Transformer layers, as the backbone network.
Following~\cite{peng22c_interspeech,10446062}, we randomly re-initialized the last three Transformer layers.
Both projector and predictor are 2-layer multi-layer perceptrons, each comprising a linear layer with an output size of 2048 followed by batch normalization~\cite{pmlr-v37-ioffe15}, GELU~\cite{hendrycks2023gaussian} activation function, and another linear layer with an output size of 256.
We optimized the student with AdamW~\cite{loshchilov2018decoupled} for 15 epochs ($\approx$ 58.6k steps) with a batch size of 6 minutes of speech segments.
The learning rate was linearly warmed up from 1e-5 to 1e-4 for 3\% of the steps, held for 47\% of the steps, and linearly decayed to 1e-5 for remaining 50\%.
During warm-up, only the randomly initialized parameters were updated.
The momentum of EMA was set to 0.999.
For speaker perturbation, we set the parameters (formant shift ratio, new pitch median, pitch range factor) to (1.1, 300, 1.2) and (1/1.1, 100, 1/1.2) for male-to-female and female-to-male conversions, respectively.
The pitch threshold for classifying conversion types was set to 155 Hz.
Regarding the minimum cut algorithm, we set the second-per-syllable to 0.2 and the merge threshold to 0.3, as recommended in the official implementation\footnotemark{}.
The numbers of clusters for $K$-Means and agglomerative clustering were 16384 and 4096, respectively.
For evaluation, we used outputs from the 8-th Transformer layer of the student based on the results in Fig.~\ref{fig:layer-wise}.

\section{RESULTS}
\label{sec:results}

\subsection{Syllable segmentation and discovered unit quality}

\begin{table*}[t]
\centering
\caption{Syllable segmentation scores and discovered syllabic unit quality on Librispeech. All scores except for mutual information are reported in percentage units [\%].}
\label{tab:segmentation}
\begin{tabular}{|l|cccc|ccc|}\hline
\multirow{2}{*}{Model}&\multicolumn{4}{|c|}{Syllable segmentation scores}&\multicolumn{3}{|c|}{Discovered syllabic unit quality}\\\cline{2-8}
&Precision&Recall&F1&R-value&Syllable purity&Cluster purity&Mutual info.\\\hline
HuBERT~\cite{9585401}&51.4&31.4&39.0&50.1&33.1&28.4&3.54\\
VG-HuBERT~\cite{peng23e_interspeech}&65.3&64.3&64.8&70.0&53.4&43.6&4.66\\
SD-HuBERT~\cite{10446062}&64.3&\textbf{71.0}&67.5&70.7&54.1&\textbf{46.2}&4.76\\
Ours (target\_layer=projector)&\textbf{73.3}&67.6&\textbf{70.3}&\textbf{74.6}&\textbf{59.4}&44.5&\textbf{5.08}\\\hline
\textit{ablation study}&&&&&&&\\
Ours--BYOL+DINO&64.3&65.1&64.7&69.8&59.1&42.9&5.06\\
Ours (target\_layer=6)&45.6&22.7&30.3&44.4&10.4&14.3&2.17\\\hline
\end{tabular}
\end{table*}

Fig.~\ref{fig:similarity_mat} shows an example of syllable boundaries obtained by the original pretrained HuBERT, SD-HuBERT, and the proposed method. In HuBERT, the similarity is mostly limited to frame-level short spans. SD-HuBERT succeeds in obtaining larger structures, but the image is noisy, having undesired frame similarity components.
The block structure is clearest in our proposed method, and the boundaries roughly match the ground truth.

Table~\ref{tab:segmentation} shows the segmentation scores and clustering quality for the Librispeech test set.
We used outputs from the 8-th Transformer layer for HuBERT, while for VG-HuBERT\footnotemark[\value{footnote}]\footnotetext{\url{https://github.com/jasonppy/syllable-discovery}} and SD-HuBERT\footnote{\url{https://github.com/cheoljun95/sdhubert}}, we employed the 9-th Transformer layer following the default configuration.
Compared to HuBERT, all VG-HuBERT, SD-HuBERT, and our proposed method significantly improved all metrics due to the emergence of syllabic organization.
Since SD-HuBERT oversegmented utterances compared to ground truth syllable boundaries, it achieved the best recall at the expense of precision.
This oversegmentation makes segment-wise averaged features purer but lowers the purity of syllables and the mutual information.
Among the three self-supervised fine-tuning methods, as a whole, our proposed method performs the best, demonstrating its efficacy.

To validate the effectiveness of two measures for speaker disentanglement, speaker perturbation and frame-level loss, we also investigated using DINO instead of BYOL in our proposed method (Ours--BYOL+DINO).
Unlike SD-HuBERT, we did not append the CLS token to Transformer encoder inputs and computed frame-wise cross-entropy loss with 4096 classes, matching the number of syllabic unit clusters.
In our DINO variant, the quality of syllabic units was almost comparable to that achieved with BYOL.
However, we observed two undesirable trends that seem to be attributed to DINO.
First, segmentation scores moderately dropped, which is consistent with the observations in image segmentation~\cite{Caron_2021_ICCV}.
Second, most categories in the final softmax function were not activated as similarly observed in~\cite{govindarajan2023dino}.
We assume the latter is a reason for the former degradation of segmentation scores, as the same category signal is propagated to different syllable representations.
Further investigation is needed to verify this assumption.

\begin{figure}[t]
\centering
\includegraphics[width=0.8\linewidth]{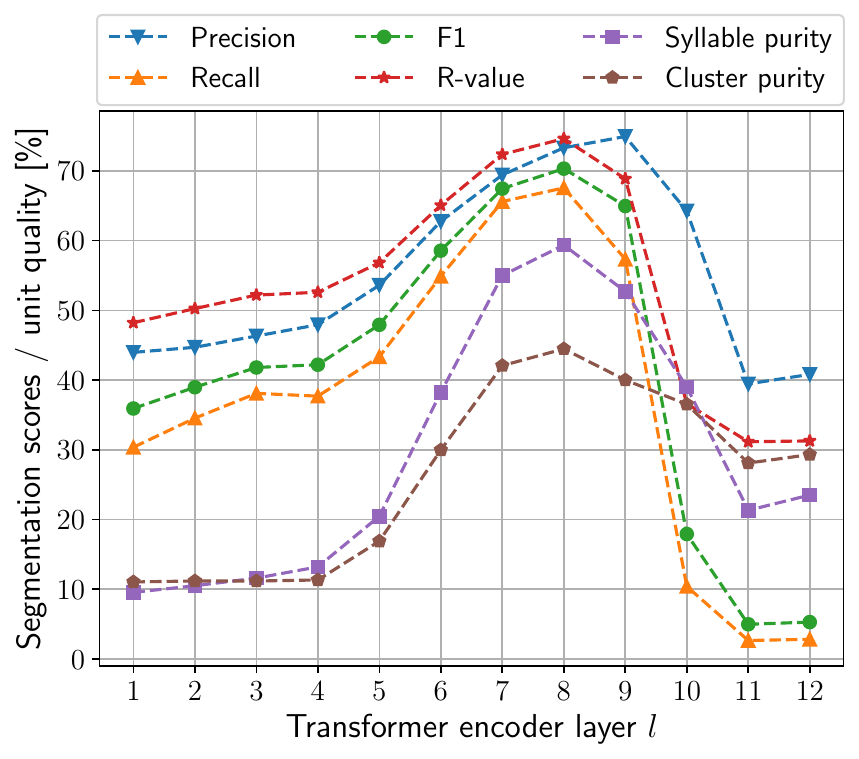}
\caption{Layer-wise analysis of syllable segmentation scores and syllabic unit quality using our proposed method.}
\label{fig:layer-wise}
\end{figure}

Fig.~\ref{fig:layer-wise} shows layer-wise analysis of syllable segmentation scores and syllabic unit quality using our proposed method. The horizontal axis indicates which layer of the student was used for evaluation. In the early layers, all scores gradually improved as the layer number increased. The F1 score peaked at the 8-th Transformer layer and subsequently decreased as frame similarity spans extended beyond the syllable boundaries.
Therefore, we selected the 8-th layer as the default setting for syllable discovery.

\subsection{Effect of speaker disentanglement}

\begin{table}[t]
\centering
\caption{Speaker identification accuracy on VoxCeleb1.}
\label{tab:sid}
\begin{tabular}{|l|c|}\hline
Model&Accuracy\\\hline
HuBERT&67.2\\
VG-HuBERT&37.4\\
SD-HuBERT&47.6\\
Ours&\textbf{26.6}\\\hline
\end{tabular}
\end{table}

Table~\ref{tab:sid} compares the speaker identification accuracies of HuBERT, VG-HuBERT, SD-HuBERT, and our method on the VoxCeleb1 test set.
We trained a linear classification head for 100 epochs ($\approx$ 430k steps) with a batch size of 256 seconds of speech segments.
The input to the linear head was sequence-averaged speech frame representations from the student's Transformer layer used for syllable segmentation.
In HuBERT and SD-HuBERT, the accuracy was relatively high because these methods do not incorporate a speaker disentanglement mechanism.
On the other hand, the accuracy of VG-HuBERT is lower due to the grounding of speech with speaker-independent visual content.
Compared to these methods, our proposed method achieved the lowest accuracy, validating the efficacy of our explicit speaker disentanglement.

\subsection{What is essential for syllabic organization?}

In Fig.~\ref{fig:similarity_mat}, we experimentally showed that syllabic organization emerges not only with a sentence-level objective but also with a frame-level objective in a speech-only model.
This finding is intriguing because this phenomenon is not exhibited in HuBERT, which is also trained with a frame-level objective.
We believe that the crucial factor for syllabic organization is the use of coarse-level linguistic information as prediction targets for the student.
While the prediction targets of HuBERT-base were mel-frequency cepstral coefficients (MFCCs) or its 6-th Transformer layer outputs, which have been shown to correlate with phones~\cite{9585401}, our method employs the projections of outputs from the last teacher Transformer layer.
Indeed, a recent study has shown that representations from the latter Transformer layers in HuBERT correlate well with words~\cite{pasad-etal-2024-self}.
Furthermore, both VG-HuBERT and SD-HuBERT predict more abstract sentence-level content.
To verify our hypothesis, we conducted an ablation study using the outputs from the teacher's 6-th Transformer layer as prediction targets for the student during training (target\_layer=6).
We set the output size of the predictor to 768, which is the hidden size of HuBERT-base, and used outputs from the student's 8-th Transformer layer for evaluation.
In Table~\ref{tab:segmentation}, we had very poor results in this setting, concluding that linguistically coarse-grained prediction targets are responsible for emergence of syllabic organization in self-supervised speech models.

\section{CONCLUSION}
We proposed a speech-only, self-supervised fine-tuning method designed to discover syllabic units while disentangling speaker information.
To extract speaker-invariant speech representations, our approach learns to match representations of the original speech with speaker-perturbed one.
Moreover, to prevent the CLS token from aggregating speaker information in SD-HuBERT, we employ a frame-level training loss without relying on the CLS token.
Experimental results showed that our speaker disentanglement method improved the quality of syllabic units.
Furthermore, we demonstrated that the essential factor for syllabic organization is the use of linguistically coarse-grained prediction targets for the student.
The speaker identification analysis confirmed that our method successfully normalized the speaker information with the help of pitch and formant frequency perturbation, achieving the lowest accuracy compared to existing methods.
Disentangling other prosody components, such as speed or rhythm, might further reduce undesired speaker dependencies and improve the quality of syllabic content representation.

\bibliographystyle{IEEEbib}
\bibliography{refs}

\end{document}